\documentclass[conference]{IEEEtran}

\IEEEoverridecommandlockouts
\usepackage{cite}
\usepackage{amsmath,amssymb,amsfonts}
\usepackage{algorithm}
\usepackage{algorithmic}
\usepackage{graphicx}
\usepackage{textcomp}
\usepackage{xcolor}
\usepackage{bm}

\usepackage{hyperref}

\usepackage{multirow}

\def\BibTeX{{\rm B\kern-.05em{\sc i\kern-.025em b}\kern-.08em
    T\kern-.1667em\lower.7ex\hbox{E}\kern-.125emX}}

\def\HU{\textsuperscript{1}}
\def\CHU{\textsuperscript{2}}
\def\NU{\textsuperscript{3}}

\begin{document}

\title{Improved Activity Forecasting\\ for Generating Trajectories}

\author{
Daisuke Ogawa\HU,
Toru Tamaki\HU,
Tsubasa Hirakawa\CHU,
Bisser Raytchev\HU,
Kazufumi Kaneda\HU,
Ken Yoda\NU\\
\HU Hirohsima University, Japan, \CHU Chubu University, Japan, \NU Nagoya Univerisity, Japan
}

\maketitle

\begin{abstract}
An efficient inverse reinforcement learning for generating trajectories is proposed
based of 2D and 3D activity forecasting.
We modify reward function with $L_p$ norm and propose convolution into value iteration steps, which is called convolutional value iteration.
Experimental results with seabird trajectories (43 for training and 10 for test),
our method is best in terms of MHD error and performs fastest.
Generated trajectories for interpolating missing parts of trajectories
look much similar to real seabird trajectories than those by the previous works.
\end{abstract}


\section{Introduction}

Analyzing, understanding and predicting human and animal movement in forms of trajectory is an
important task \cite{Morris2011TPAMI,Li2015survey,DBLP:conf/hci/HirakawaYTF18} and has been studied in computer vision
\cite{MDA,Ogawa2018FCV} as well as ecology \cite{Hirakawa2017,hirakawa2018can}.
Despite of recent recent progress in deep learning
\cite{yi2016pedestrian,alahi2016social,fernando2018soft+,DBLP:journals/corr/LeeCVCTC17},
those methods are hard to apply to ecological data because of small size of datasets; at most few dozens of trajectories are available because it is not an easy task to obtain trajectory data of animals.

In this paper we propose an improvement of activity forecasting \cite{kitani2012activity} that predicts distributions of trajectories over a 2D plane by using maximum entropy (MaxEnt) based \cite{ziebart2008maximum} inverse reinforcement learning (IRL) \cite{ng2000algorithms}. Activity forecasting has been applied to trajectories of birds by extending the state space from 2D to 3D, in order to simulate bird behavior by generating trajectories \cite{Hirakawa2017} and interpolate missing parts in trajectories \cite{hirakawa2018can}.
However the computation cost in both time and space is very demanding because of the 3D extension.
We investigate formulations and implementations of previous works, derive an improvement and efficient formulation,
and present a stochastic method for generating trajectories that are more similar to real trajectories than those by the previous deterministic approaches.

\section{Methods}

\subsection{Reinforcement and inverse reinforcement learning}

Reinforcement learning (RL) \cite{kaelbling1996reinforcement,sutton2018reinforcement}
formulates a problem in which agents in an environment take actions for maximizing rewards,
and modeled by Markov Decision Process (MDP).
The agent gains immediate reward $r_t$ at state $s_t$ by choosing action $a_t$ at time step $t$, then move to the next state according to state transition probability $p$.
Through a learning procedure, we obtain a policy $\pi$ that maximizes the accumulated reward $R$ through a series of actions.
For a small discrete problem in 2D, the state space is usually discretized as a grid, and a trajectory is
represented by a series (or trajectory)  $\xi = \{s_1, \ldots, s_t\}$ of 2D grid coordinates $s_t = (x_t, y_t)$,
where $x_t$ and $y_t$ are two-dimensional grid coordinates. 
Actions $a_t$ can be defined as moves to neighboring eight grid states.

A common approach to RL is Q-leaning \cite{sutton2018reinforcement}
that iteratively update action value function $Q(s_t, a_t)$ with weight $\alpha$ and discount factor $\gamma$
by
\begin{equation}
Q(s_t, a_t) \gets (1 - \alpha)Q(s_t, a_t) + \alpha(r_{t+1} + \gamma\max_{a^{\prime}} Q(s_{t+1}, a^{\prime}) ),
\label{eq:value_iteration}
\end{equation}
that approximates the Bellman equation after convergence.
Then the greedy policy taking the action that maximizes $Q$ at $s_t$, is optimal.

Reward values are however not always given or difficult to define in practice,
therefore inverse Reinforcement Learning (IRL) \cite{ng2000algorithms} is used to estimate the reward values,
and the maximum entropy (MaxEnt) approach \cite{ziebart2008maximum} formulates the IRL problem as follows.

\subsection{MaxEnt IRL}

The reward value of trajectory $\xi$ is expressed by linear combination of feature vector $\bm{f}$,
\begin{equation}
R(\xi; \bm{\theta}) = \sum_t r(s_t; \bm{\theta}) = \sum_t \bm{\theta}^{T} \bm{f}(s_t),
\end{equation}
where $\bm{\theta}$ is the parameter to be estimated,
and $\bm{f}(s_t)$ is a pre-defined feature vector value at $s_t$.
This approach defines probability of trajectory $\xi$ by\footnote{The denominator (partition function) has $\theta$ and hence should be considered (but omitted here for simplicity).}
\begin{equation}
p_\pi(\xi| \bm{\theta}) 
\propto \exp(R(\xi; \bm{\theta}))
= \exp\left(
    \sum_t \bm{\theta}^{T} \bm{f}(s_t)
    \right)
\end{equation}
and solve the following log-likelihood maximization problem;
\begin{equation}
\mathop\mathrm{argmax}_{\bm{\theta}} L(\theta)
= \mathop\mathrm{argmax}_{\bm{\theta}} \frac{1}{|Z|} \sum_{\xi \in Z} \log p_\pi(\xi| \bm{\theta}),
\end{equation}
where $Z$ is a given set of trajectories.
The weight vector $\bm{\theta}$ is updated by
\begin{equation}
\bm{\theta} \gets \bm{\theta} \mathrm{e}^{\lambda \nabla_{\bm{\theta}} L(\bm{\theta})},
\end{equation}
until convergence with leaning late $\lambda$.
The gradient $\nabla_{\bm{\theta}} L(\bm{\theta})$ is given by
\begin{equation}
\nabla_{\bm{\theta}} L(\bm{\theta})
= \overline{f} - E_{p_\pi(\xi^{\prime}| \bm{\theta})} \biggl[ \sum_t  \bm{f}(s^{\prime}_t) \biggr],
\end{equation}
where $\overline{f} = \frac{1}{|Z|} \sum_{\xi \in Z}  \sum_t  \bm{f}(s_t)
%= E_{p_\pi(\xi)} \biggl[ \sum_t  \bm{f}(s_t) \biggr]
$.
Note that the expectation in the second term is taken for all possible path $\xi'$ according to policy $\pi$,
which is intractable,
and is approximated by the following weighted sum for given trajectories $Z$,
\begin{equation}
E_{p_\pi(\xi'| \bm{\theta})} \biggl[ \sum_t  \bm{f}(s_t) \biggr] 
\approx \frac{1}{|Z|} \sum_{\xi \in Z}  \sum_t  \bm{f}(s_t)D_{\xi}(s_t),
\end{equation}
where $D_{\xi}(s_t)$ is expected state frequency.

\subsection{Activity forecasting in 2D}

Activity forecasting
\cite{kitani2012activity} utilized the MaxEnt IRL approach for dealing with noisy pedestrian trajectories by
implicitly introducing hidden states. Action-value function $Q$ and state-value function $V$ are defined as follows;
\begin{align}
Q(s, a) &= r(s; \bm{\theta} ) + E_{p(s^{\prime}| s, a)}[V(s^{\prime})] \label{eq:Q}\\
V(s) &= \mathop\mathrm{softmax}_a Q(s, a), \label{eq:V}
\end{align}
where $p(s^{\prime}| s, a)$ is state transition probability, and softmax is a soft version of the maximum
that is defined as $\mathop\mathrm{softmax}_a Q(s, a) = \max_a Q(s, a) + \log[1 + \exp\{ \min_a Q(s, a) - \max_a Q(s, a)\}]$ in their implementation\footnote{\url{http://www.cs.cmu.edu/~kkitani/datasets/}}.
The policy $\pi$ is then defined by
\begin{equation}
\pi(a | s, \bm{\theta}) \propto \exp(Q(s, a) - V(s)).
\label{eq:policy}
\end{equation}
The backward-forward algorithm is shown in Algorithm \ref{alg2}
that compute the policy and $D(s)$.

\begin{algorithm}[t]
\caption{Backward-forward algorithm \cite{kitani2012activity,ziebart2008maximum}}         
\label{alg2}                          
\begin{algorithmic}
\STATE (Backward pass)
\STATE $V(s) \gets -\infty$
\FOR{$n=N$ to $1$}
\STATE $V^{(n)}(s_{goal}) \gets 0$
\STATE $Q^{(n)}(s, a) = r(s; \bm{\theta} ) + E_{p(s^{\prime}| s, a)}[V(s^{\prime})]$
\STATE $V^{(n-1)}(s) = \mathop\mathrm{softmax}_a Q^{(n)}(s, a)$
\ENDFOR
\STATE $\pi_{\bm{\theta}}(a | s) \propto \exp(Q(s, a) - V(s))$
\STATE 
\STATE (Forward pass)
\STATE $D(s_{initial}) \gets 1$
\FOR{$n=1$ to $N$}
\STATE $D^{(n)}(s_{goal}) \gets 0$
\STATE $D^{(n+1)}(s) = \sum_{s^{\prime}, a} p(s^{\prime}| s, a) \pi_{\bm{\theta}}(a | s^{\prime}) D^{(n)}(s^{\prime})$
\ENDFOR
\STATE $D(s) = \sum_{n}D^{(n)}(s)$
\end{algorithmic}
\end{algorithm}

\subsection{Activity forecasting in 3D}

Hirakawa et al. \cite{hirakawa2018can} extended the 2D activity forecasting to 3D for dealing with time explicitly to achieve accurate interpolation of missing parts in GPS trajectories of seabirds.
They extend two-dimensional states to three-dimensional states by augmenting time step;
a state is defined as $s_t = (x_t, y_t, z_t)$, where $x_t$ and $y_t$ are 2D grid coordinates and $z_t$ is a discrete time step.
This is because seabirds may take an indirect route between two points, while IRL methods including 2D activity forecasting tend to produce direct routes.
In this work, an action is defined as $a = (a_{xy}, 1)$, where $a_{xy}$ is moves to neighboring eight grid states in 2D,
and the last augmented value of one enforces the increment in time step when taking one action.
Now a trajectory is denoted by $\xi = \{(s_0, a_0), (s_1, a_1), \dots \}$.
Because of this 3D extension, their 3D method performs better than the original 2D forecasting,
at the cost of increased computation time and required memory.
This cost is very demanding and therefore efficient improvements would be necessary.

% MHD and computation cost table
\begin{table*}[t]
\caption{MHD of interpolation results and computation time. Average values with std are reported.
$p=2$ or $p=3$ indicates $L_p$ norm used in the modified reward function.
w/ or w/o indicates convolutional or ordinal value iteration.
Note that computation time of value iteration is for a single iteration of Eqs.(\ref{eq:Q}) and (\ref{eq:V}),
and update of $\theta$ is for computation of $\nabla_\theta L(\theta)$ for a single update.
Columns of ratio is based on row ''w/o conv''.
}
\centering
\begin{tabular}{c|cc|cc|cc}
                      & MHD (deterministic)
                      & MHD (stochastic)
                      & value iteration [s]
                      & ratio 
                      & update of $\theta$ [s]
                      & ratio        
                      \\ \hline
Linear  & $12.20 \pm 4.48$ &  &   &   &   &   \\
3D \cite{hirakawa2018can}
& $5.33 \pm 2.61$
&
& $4.0 \pm 0.19$
& $1191$
& $22354 \pm 449$
& $2064$ \\
2D \cite{kitani2012activity}
& $5.80 \pm 2.96$
& $6.02 \pm 2.97$
& $0.008667 \pm 0.003229$
& $2.57$
& $15.05 \pm 0.78$
& $1.39$ \\
$p=2$ w/o conv
& $6.37 \pm 3.19$
& \bm{$5.13 \pm 2.77$}
& \multirow{2}{*}{\bm{$0.003367 \pm 0.002389$}}
& \multirow{2}{*}{$1$}
& \multirow{2}{*}{\bm{$10.83 \pm 0.85$}}
& \multirow{2}{*}{$1$} \\
$p=3$ w/o conv
& $6.14 \pm 3.35$
& $5.20 \pm 3.09$
&
&
&
&
\\
$p=2$ w/ conv
& $6.45 \pm 3.36$
& $5.22 \pm 2.82$
& \multirow{2}{*}{$0.003542 \pm 0.002242$}
& \multirow{2}{*}{$1.05$}
& \multirow{2}{*}{$11.5608 \pm 0.8293$}
& \multirow{2}{*}{$1.07$} \\
$p=3$ w/ conv 
& $5.63 \pm 3.32$
& $5.20 \pm 3.06$
&
&
&
&
\end{tabular}
\label{tab:result}
\end{table*}

\subsection{Proposed method}

To achieve a better performance with a smaller computation cost, we propose the following improvements.

First, we take 2D approach like as the original activity forecasting \cite{kitani2012activity}.
A 3D extension approach \cite{hirakawa2018can}
may work better, but inherently increases computation cost because of three-dimensional state space.

Second, we ''exactly'' follow the definitions of $Q$ and $V$ as shown in Eqs. (\ref{eq:Q}) and (\ref{eq:V}).
Implementations of previous works\footnote{
\url{http://www.cs.cmu.edu/~kkitani/forecasting/code/oc.cpp} for \cite{kitani2012activity}
and 
\url{https://github.com/thirakawa/MaxEnt_IRL_trajectory_interpolation} for \cite{hirakawa2018can}.
}
are different from the equations, and evaluate the softmax function for eight actions.
This is one of the most computationally intensive part in the implementations.
The use of softmax is introduced in \cite{ziebart2008maximum}, but in preliminary experiments
we found that softmax doesn't affect results so much and can be replaced with max as in common value iteration.

Third, we make reward values dependent to next state $s'$ as well as current state $s$
to imitate the effect of the original implementation of 2D activity forecasting.
It is natural to move every direction in 2D with the same cost under the same condition,
however the eight moves in neighboring 2D grid state result in the preference to diagonal moves.
For example, one step to north-west from the current state is longer than a step to the west,
but the reward is a function of current state, $r(s)$.
Therefore we propose to define the reward as $r(s, s', a)$ in order to take the distance between adjacent states.
More speficically,
\begin{align}
    r(s, s', a) = r(s) / \mathrm{dist}_p (s, s'),
\end{align}
where $\mathrm{dist}_p (s, s')$ is $L_p$ distance between states $s$ and $s'$.
In experiments, we compare results with $p=2$ and $p=3$ because $p=2$ is the Euclidean distance, which seems to be natural in this task, and $p=3$ produces results similar to the effect by the original 2D activity forecasting implementation.

Fourth, we propose an effective value iteration called \textit{convolutional value iteration}.
The core of value iteration is to iterate Eq.(\ref{eq:Q}) and Eq.(\ref{eq:V}) to propagate values.
Here, we insert a convolution step with the Gaussian kernel $G$ to make $V$ blur, which effectively accelerate
the propagation and convergence of iteration.
The proposed method has the following form of iteration;
\begin{align}
Q(s, a) &= E_{p(s^{\prime}| s, a)}[r(s, s', a; \bm{\theta} ) + V(s^{\prime})] \label{eq:Q_mod}\\
V(s) &= (\max_a Q(s, a)) \otimes G \label{eq:V_mod}
\end{align}

Fifth, we define policy $\pi$ by
\begin{equation}
\pi(a | s, \bm{\theta}) \propto \exp(Q(s, a)),
\label{eq:policy_mod}
\end{equation}
which is more common in RL than Eq.(\ref{eq:policy}).

\subsection{Trajectory generation}

Once a policy has been obtained, a trajectory can be generated either deterministic or stochastic way.
A deterministic trajectory interpolation is to obtain next states by repeatedly selecting the action that
maximize the policy at current states.
A stochastic interpolation instead selects an action by sampling according to the policy.
We compare two ways; \cite{hirakawa2018can} generated deterministic trajectories, however those
look rather straight compared to real trajectories. Stochastic trajectories are expected to be more realistic.

\section{Experimental results}

The experimental setting is the same with \cite{hirakawa2018can};
we used 43 trajectories for training and 10 trajectories for test.
and each test trajectory has missing parts to be interpolated.
We compared our proposed method with linear interpolation,
activity forecasting in 2D \cite{kitani2012activity} and 3D\cite{hirakawa2018can}.

The modified Hausdorff distance (MHD) \cite{dubuisson1994modified}
was used as a metric for quantitative evaluation.
MHD is used to compute the similarity between object shapes.
Given two trajectories, or two series of points $\bm{A} = \{a_1, \dots, a_{N_a}\}$ and $\bm{B} = \{b_1, \dots, b_{N_b}\}$,
MHD between $\bm{A}$ and $\bm{B}$ is defined by
\begin{equation}
\mathrm{MHD}(\bm{A}, \bm{B}) = \max\left\{\frac{1}{N_a}\sum_{a \in \bm{A}}d(a, \bm{B}), \frac{1}{N_b}\sum_{b \in \bm{B}}d(b, \bm{A})\right\},
\end{equation}
where $d(a, \bm{B}) = \min_{b \in \bm{B}}||a - b||$.

Table \ref{tab:result} shows experimental results.
In terms of MHD,
results of our proposed method with $p=2$ or $p=3$ with or without convolutional value iteration
are not better than the previous works when deterministic trajectories are used.
However with stochastic trajectory generation, our method works better than 2D activity forecasting.
Convolutional value iteration was not observed to be effective for producing stochastic trajectory generation,
while it may help for a faster convergence.
Note that we exclude results of stochastic version of 3D approach because it doesn't guarantee to generate trajectories ending at the given time with stochastic policy sampling.

In terms of computation cost, 
our method performs much faster than 3D approach \cite{hirakawa2018can}, which is more than factor of 1000,
and even faster than the original 2D approach \cite{kitani2012activity}.
Note that all implementations were written in python and evaluated on the same computer.
Required memory by our method is also much smaller than 3D approach, and almost similar to 2D approach (not reported here).

\begin{figure}[tb]
\centering
\includegraphics[keepaspectratio, width=0.8\linewidth]{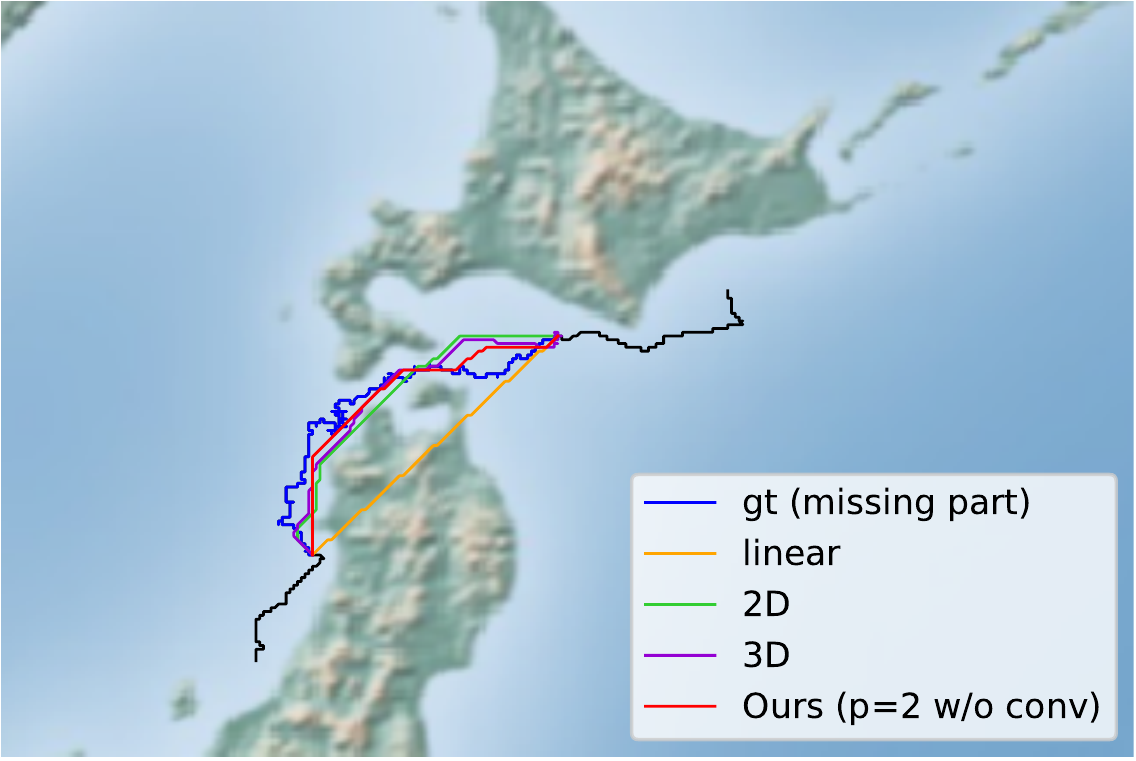}
\caption{Deterministic interpolation of the missing part in a trajectory by the proposed method and baseline methods. ``gt'' is ground-truth of the trajectory.}
\label{fig:vs-all-det}
\end{figure}

\begin{figure}[tb]
    \centering
        \includegraphics[keepaspectratio, width=0.6\linewidth]{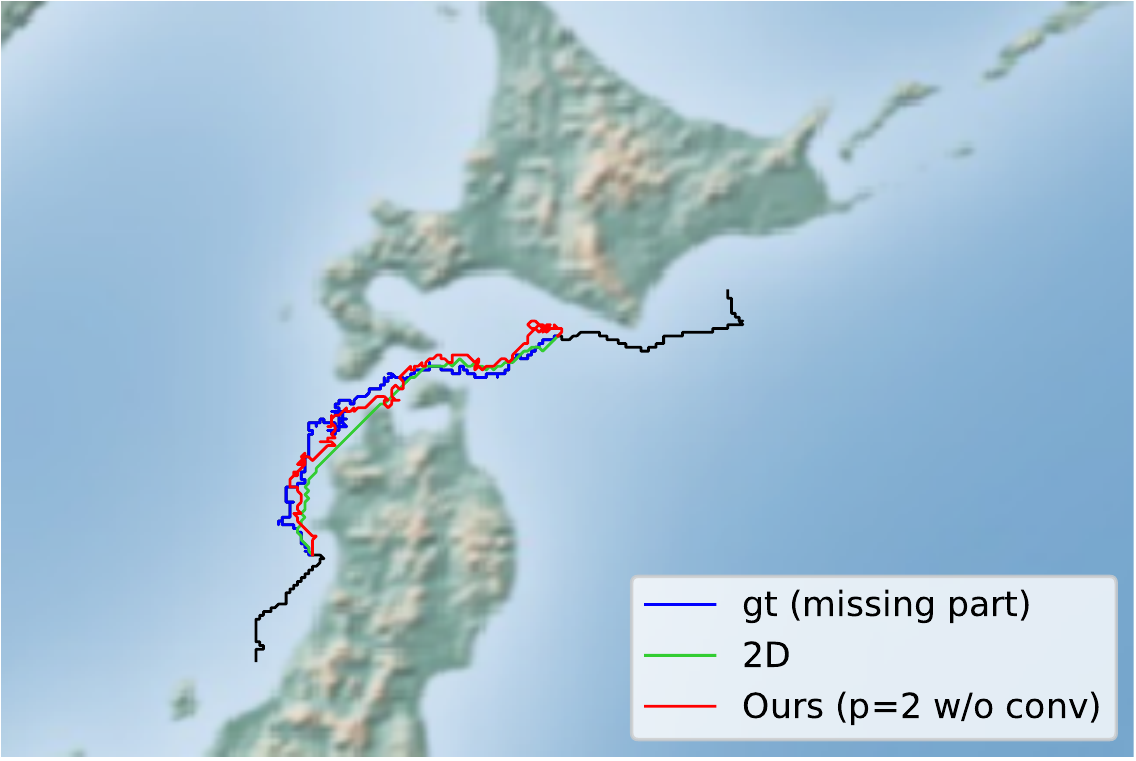}
        \includegraphics[keepaspectratio, width=0.6\linewidth]{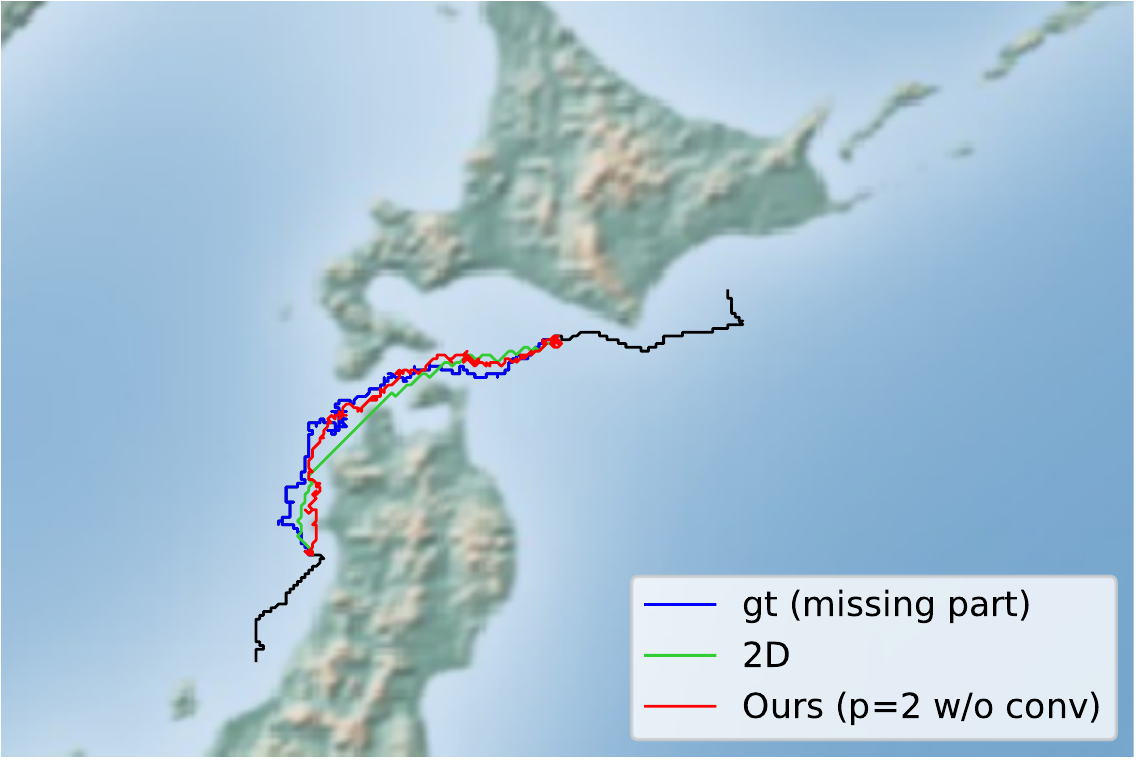}
        \includegraphics[keepaspectratio, width=0.6\linewidth]{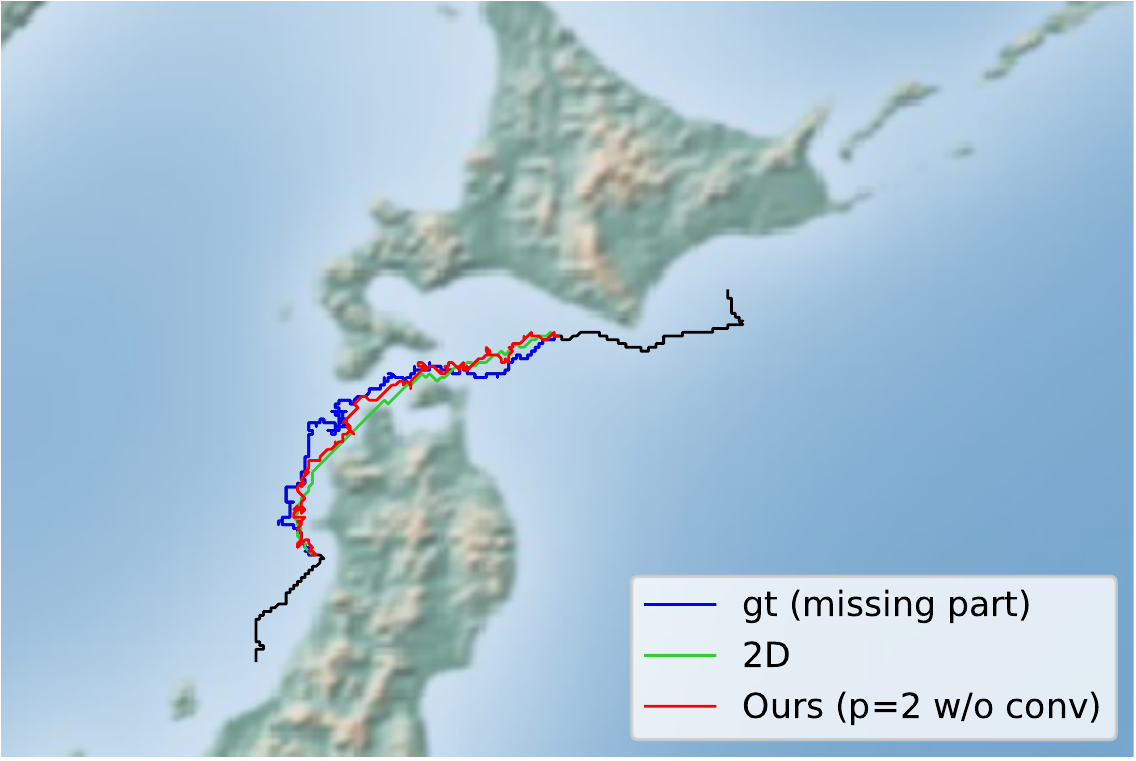}
        \includegraphics[keepaspectratio, width=0.6\linewidth]{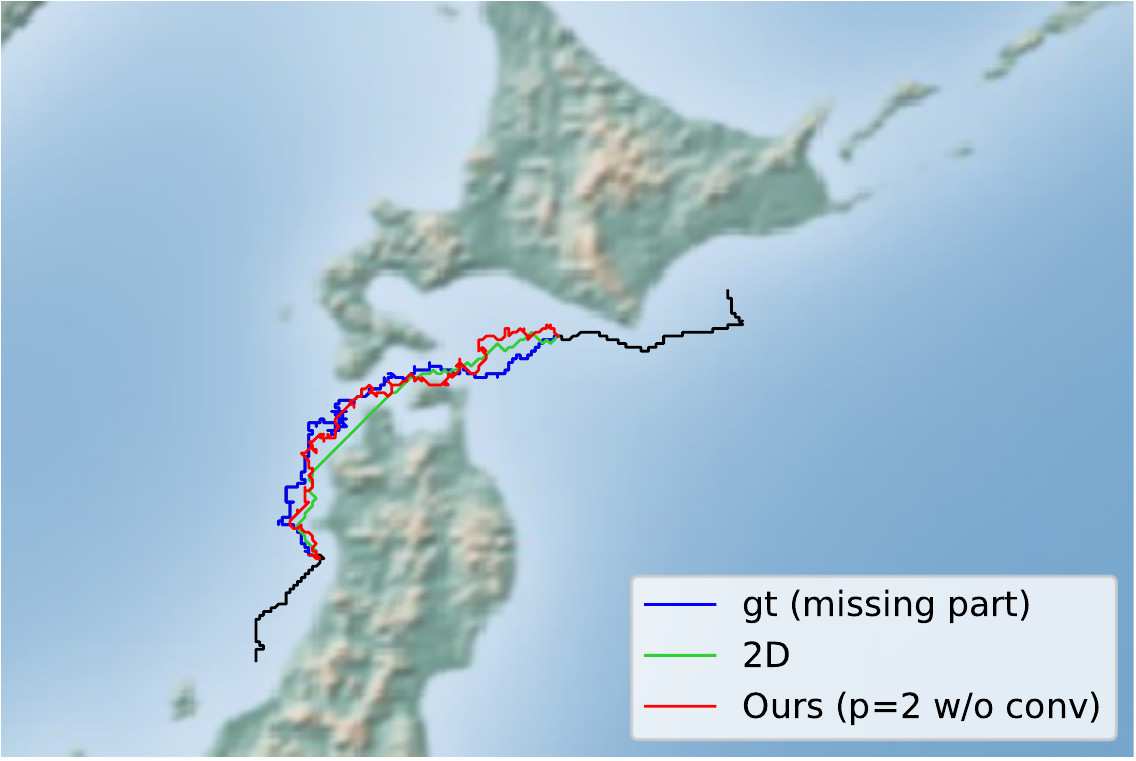}
        \includegraphics[keepaspectratio, width=0.6\linewidth]{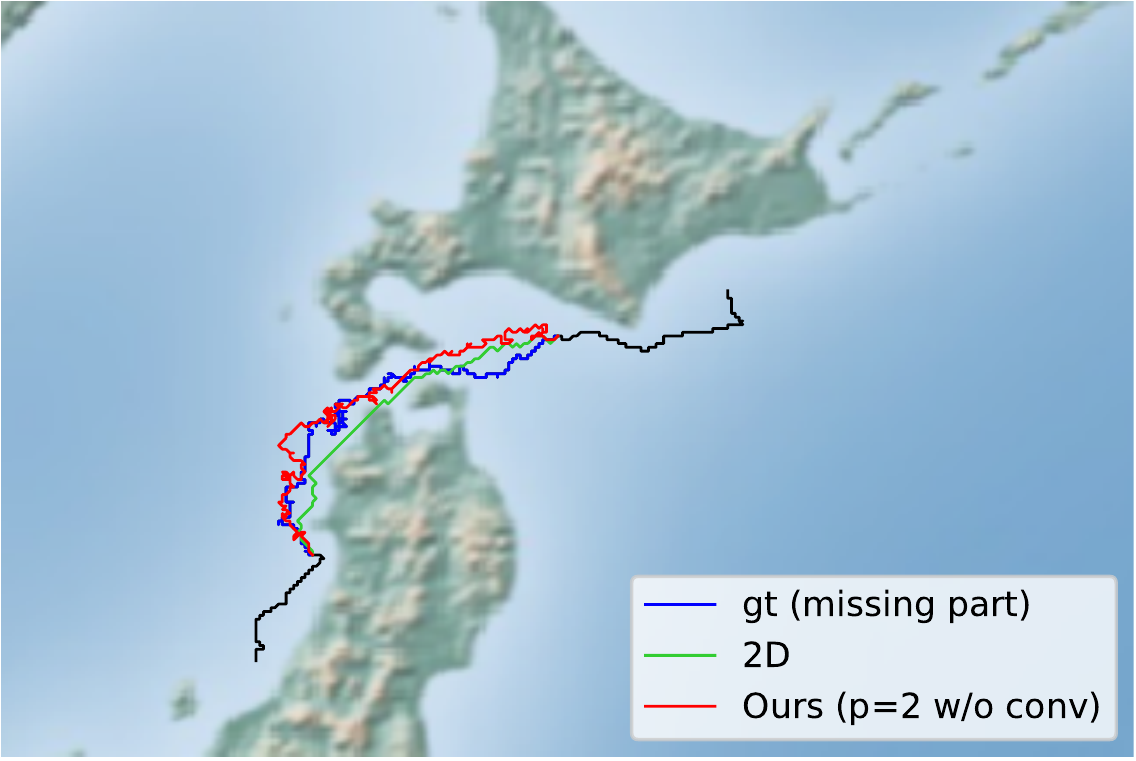}
\caption{Five different results of stochastic interpolation of the missing part in a trajectory by the proposed method and 2D approach. ``gt'' is ground-truth of the trajectory.}
\label{fig:vs-sto}
\end{figure}

Figure \ref{fig:vs-all-det} shows an example of deterministic trajectory interpolation with different methods.
Due to the nature of deterministic trajectory generation, results by all methods look similar and straight vertically, horizontally, or diagonally.
Figure \ref{fig:vs-sto} shows stochastic interpolation results of five different runs.
Results of 2D and 3D approaches still looks straight, which may caused by estimated policies that assign larger probabilities to one single action at each state.
In comparison, our method succeeded to generate more realistic trajectories. This might be attributed to
the modified reward and policy.

\begin{figure}[tb]
\centering
\includegraphics[keepaspectratio, width=0.8\linewidth]{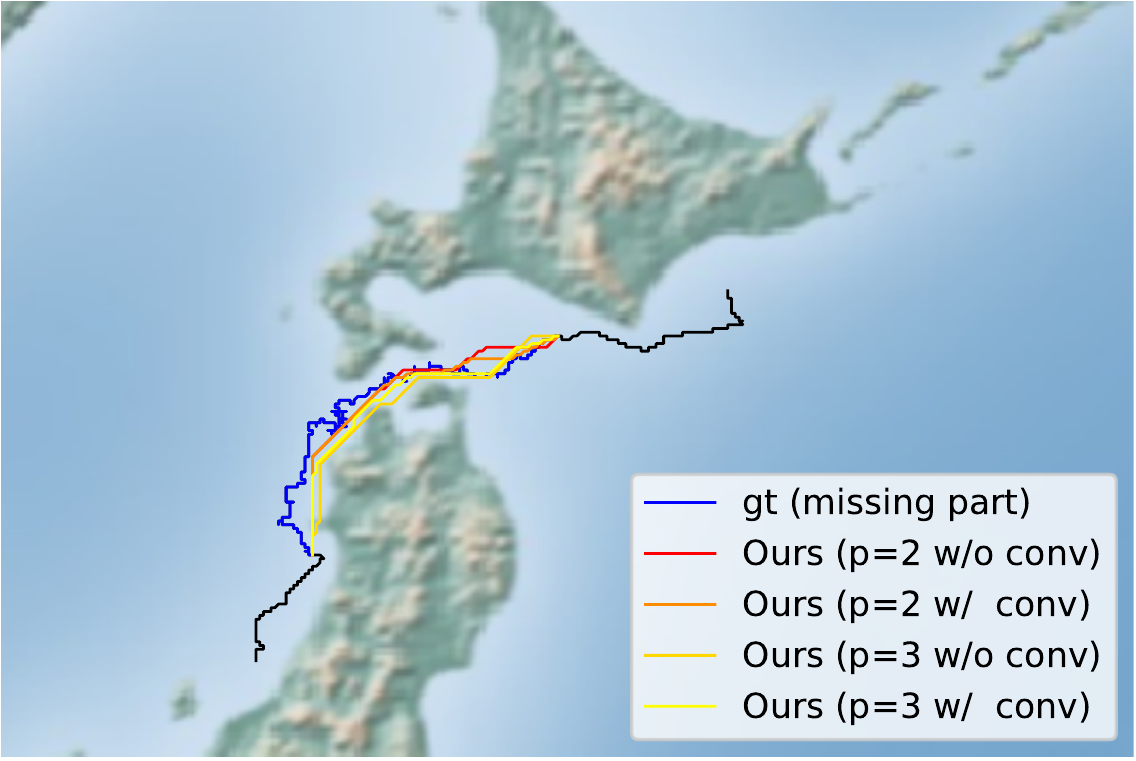}
\caption{Deterministic interpolation of the missing part in a trajectory by the proposed method with different parameters. ``gt'' is ground-truth of the trajectory.}
\label{fig:vs-2d-det}
\end{figure}

\begin{figure}[tb]
        \centering
        \includegraphics[keepaspectratio, width=0.6\linewidth]{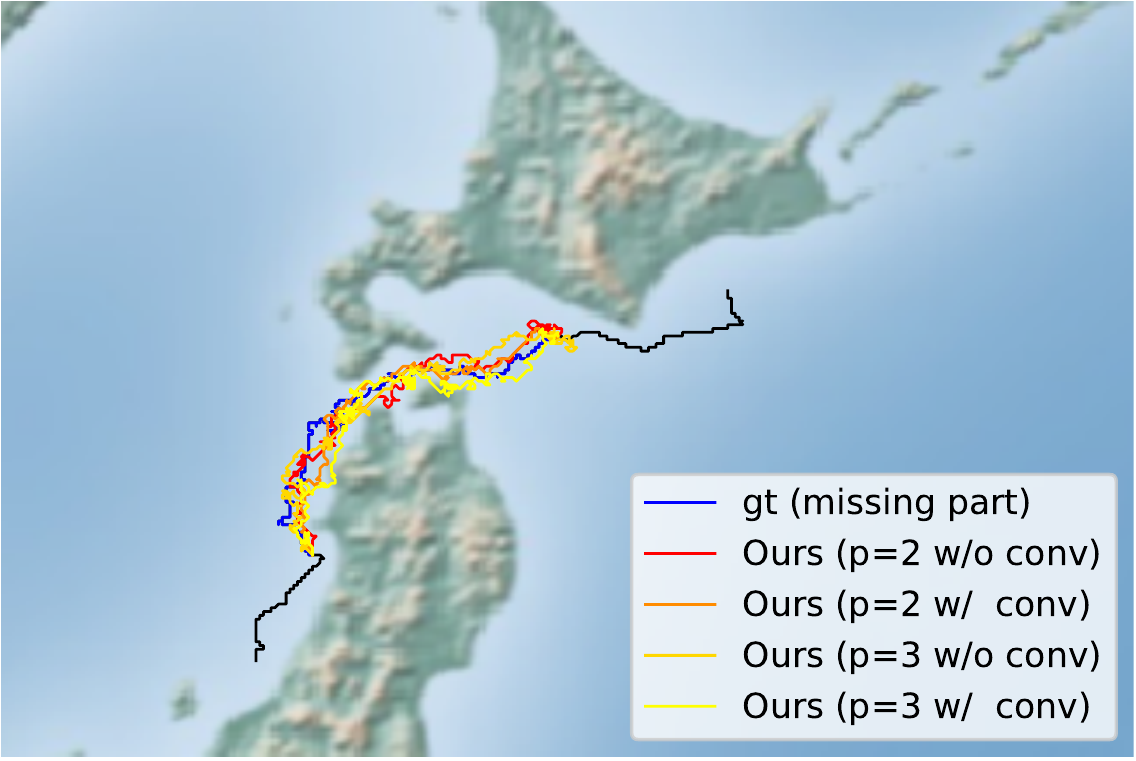}
        \includegraphics[keepaspectratio, width=0.6\linewidth]{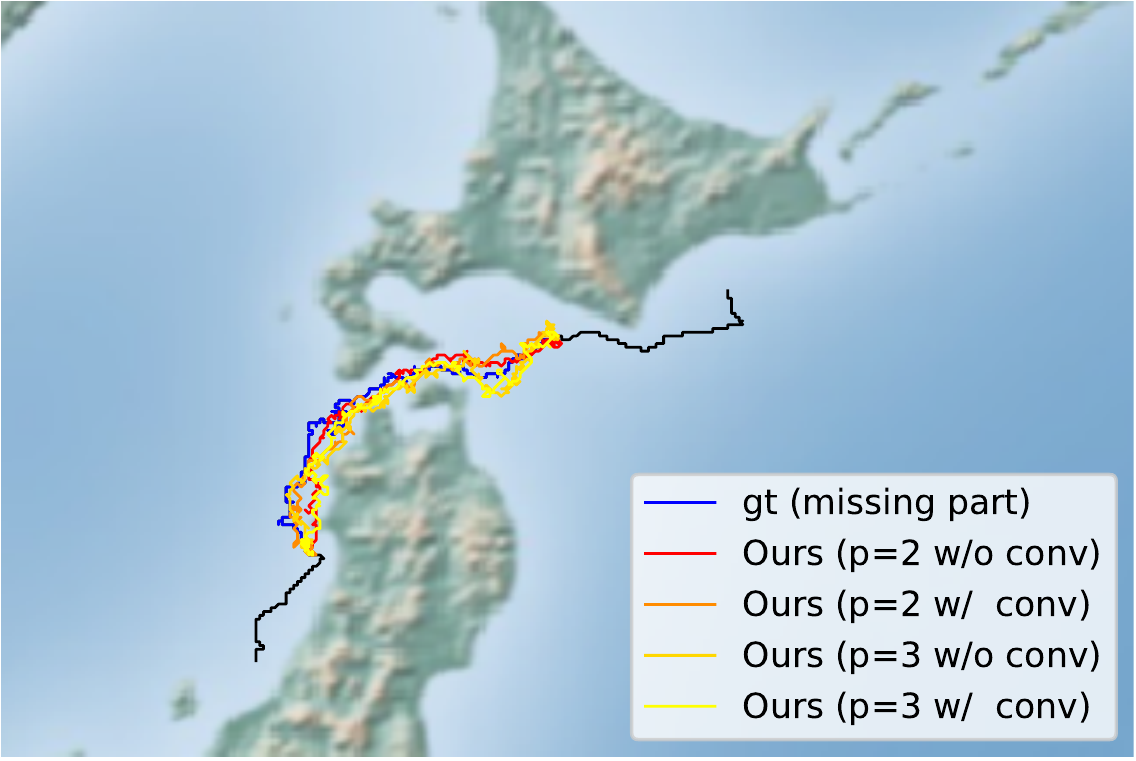}
        \includegraphics[keepaspectratio, width=0.6\linewidth]{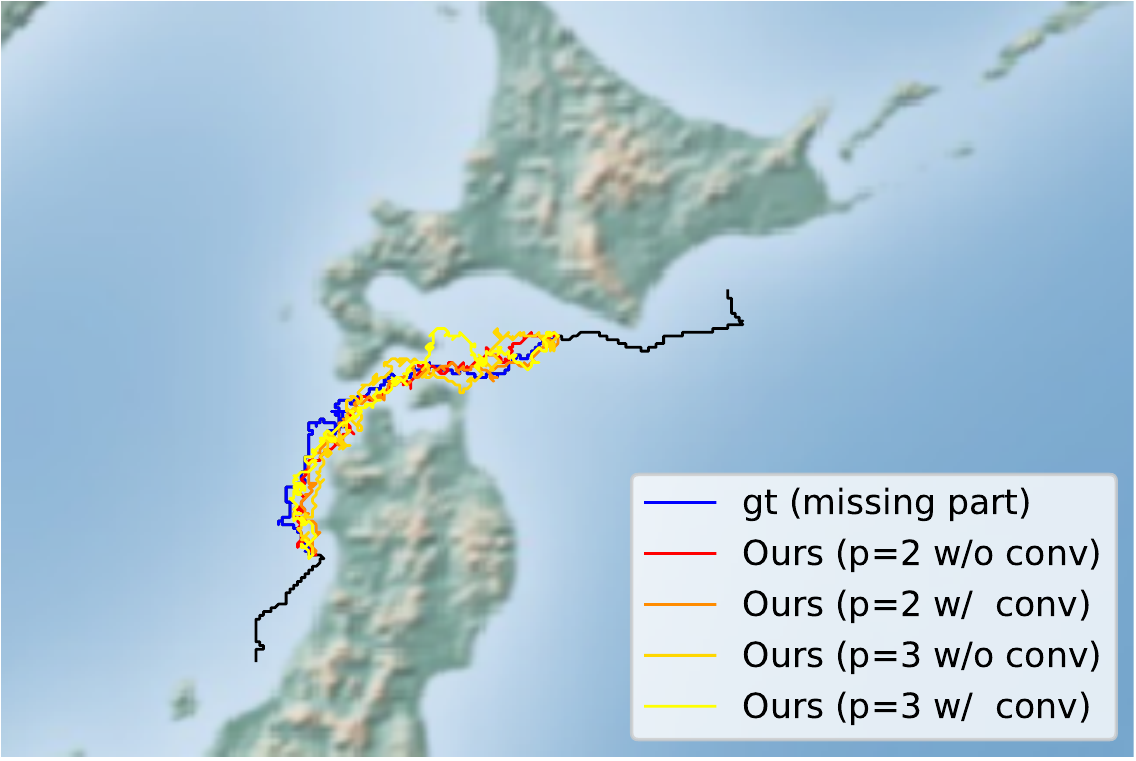}
        \includegraphics[keepaspectratio, width=0.6\linewidth]{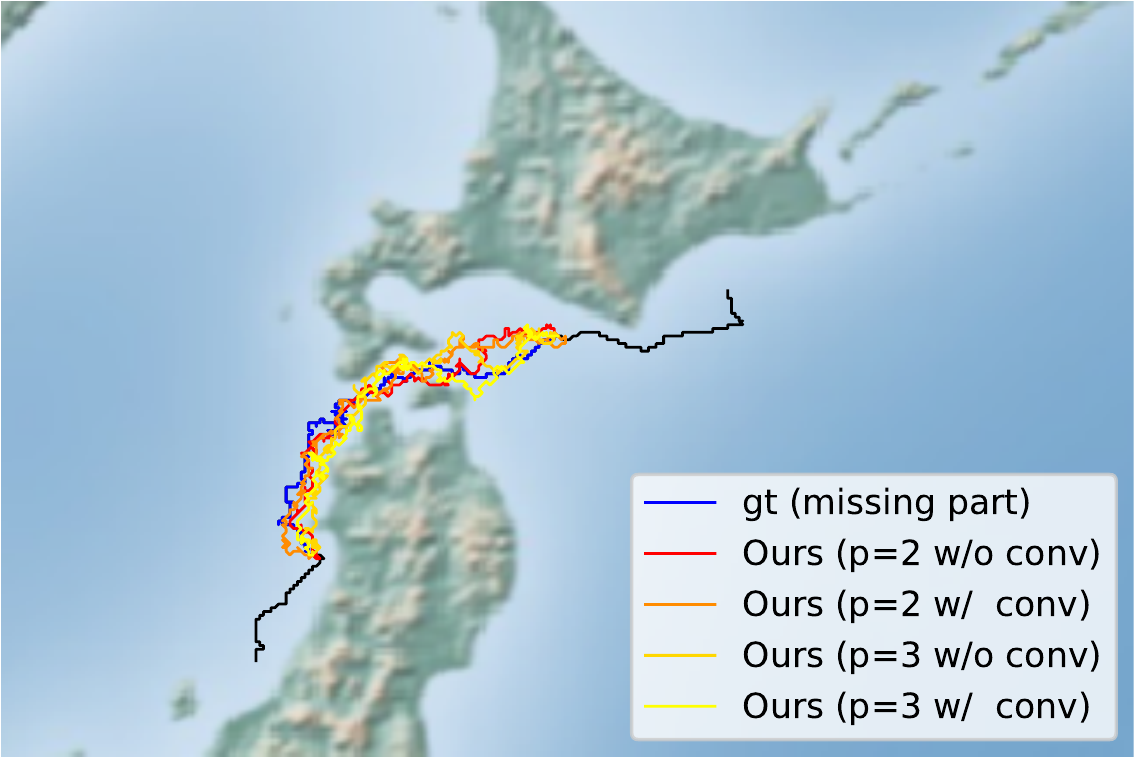}
        \includegraphics[keepaspectratio, width=0.6\linewidth]{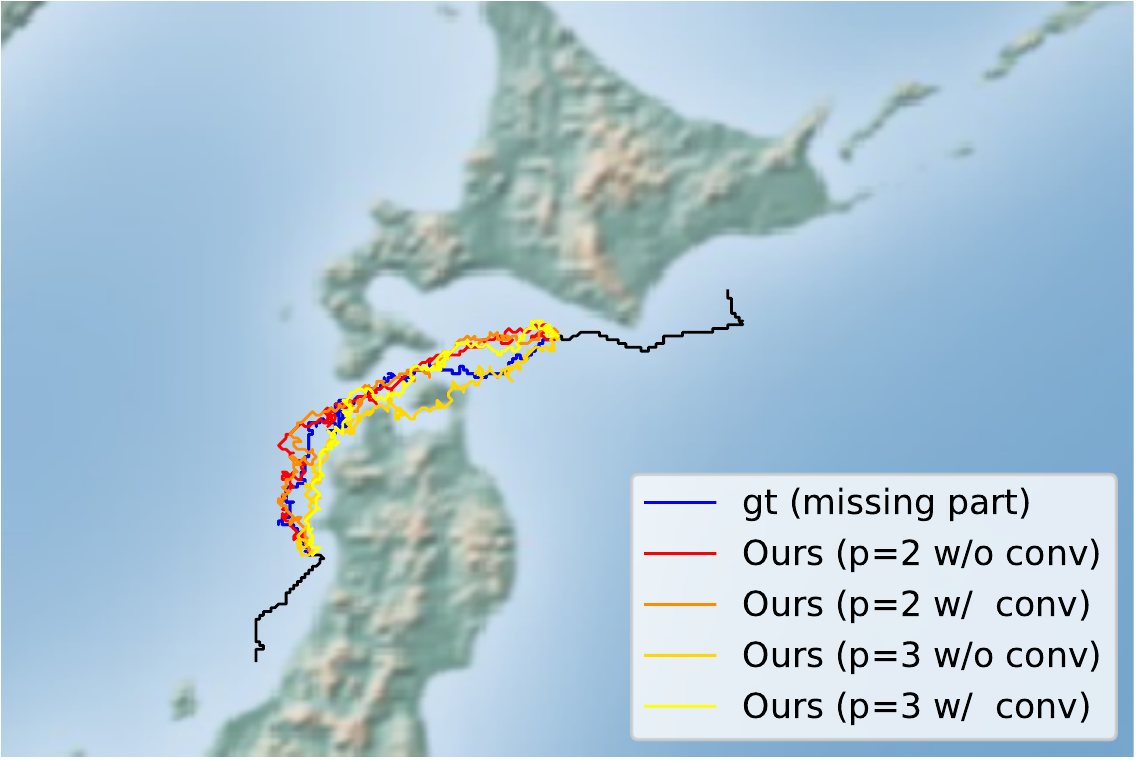}
\caption{Five different results of stochastic interpolation of the missing part in a trajectory by the proposed method with different parameters. ``gt'' is ground-truth of the trajectory.}
\label{fig:vs-2d-sto}
\end{figure}

Figures \ref{fig:vs-2d-det} and \ref{fig:vs-2d-sto}
show results of our method with different parameters.
Interpolation results look similar to each other and therefore parameters doesn't affect results.

% vs-all-sto

% vs-2d-sto

\section{Conclusion}
We have proposed an efficient inverse reinforcement learning for generating trajectories based on 2D and 3D activity forecasting.
Experimental results with a real dataset demonstrated that the proposed method works faster than 3D approach and effective for generating realistic trajectories.
Future work includes handling temporal information. We have chosen a 2D approach, however time is still an important factor for interpolation and generation of trajectories. Keeping computation cost small and adding temporal factor is an challenging problem.

\section*{Acknowledgments}
This work was supported by JSPS KAKENHI grant numbers
JP16H06540, % tamaki
JP16K21735, % kokusai
and
JP16H06541. % yoda

\bibliographystyle{unsrt}
\bibliography{mybib}

\end{document}